\documentclass[letterpaper, 10 pt, conference]{ieeeconf}  

\IEEEoverridecommandlockouts                              

\usepackage{graphics} 
\usepackage[pdftex]{graphicx}   
\usepackage{subcaption}         
\usepackage{epsfig} 
\usepackage{mathptmx} 
\usepackage{times} 
\usepackage{amsmath} 
\usepackage{amssymb}  
\usepackage{array} 
\title{\LARGE \bf
Aerial Grasping with Soft Aerial Vehicle Using Disturbance Observer-Based Model Predictive Control*
}

\author{Hiu Ching Cheung$^{1}$, Bailun Jiang$^{1}$, Yang Hu$^{1}$, Henry K. Chu$^{2}$, Chih-Yung Wen$^{1}$ and Ching-Wei Chang$^{3}$
\thanks{*This work was supported by the Research Centre of Unmanned Autonomous Systems (RCUAS) [P0046487], The Hong Kong Polytechnic University}
\thanks{$^{1}$Hiu Ching Cheung, Bailun Jiang, Yang Hu, and Chih-Yung Wen are with Department of Aeronautical and Aviation Engineering, The Hong Kong Polytechnic University, Hong Kong
        {\tt\small athena-hiu-ching.cheung@connect.polyu.hk}, {\tt\small bailun-robin.jiang@connect.polyu.hk}, {\tt\small 
        yang-h.hu@connect.polyu.hk}, {\tt\small 
        chihyung.wen@polyu.edu.hk}}%
\thanks{$^{2}$Henry K. Chu is with Department of Mechanical Engineering, The Hong Kong Polytechnic University, Hong Kong 
       {\tt\small henry.chu@polyu.edu.hk}}%
\thanks{$^{3}$Ching-Wei Chang is with The Hong Kong University of Science and Technology, Hong Kong
        {\tt\small ccw@ust.hk}}%
}

\begin{document}

\maketitle
\thispagestyle{empty}
\pagestyle{empty}

\begin{abstract}
Aerial grasping, particularly soft aerial grasping, holds significant promise for drone delivery and harvesting tasks. However, controlling UAV dynamics during aerial grasping presents considerable challenges. The increased mass during payload grasping adversely affects thrust prediction, while unpredictable environmental disturbances further complicate control efforts. In this study, our objective aims to enhance the control of the Soft Aerial Vehicle (SAV) during aerial grasping by incorporating a disturbance observer into a Nonlinear Model Predictive Control (NMPC) SAV controller. By integrating the disturbance observer into the NMPC SAV controller, we aim to compensate for dynamic model idealization and uncertainties arising from additional payloads and unpredictable disturbances. Our approach combines a disturbance observer-based NMPC with the SAV controller, effectively minimizing tracking errors and enabling precise aerial grasping along all three axes.
The proposed SAV equipped with Disturbance Observer-based Nonlinear Model Predictive Control (DOMPC) demonstrates remarkable capabilities in handling both static and non-static payloads, leading to the successful grasping of various objects. Notably, our SAV achieves an impressive payload-to-weight ratio, surpassing previous investigations in the domain of soft grasping. Using the proposed soft aerial vehicle weighing 1.002 kg, we achieve a maximum payload of 337 g by grasping.
\end{abstract}

\begin{keywords}
Aerial Grasping, Disturbance Observer, Extended Kalman Filter, Model Predictive Control, Soft Aerial Vehicle
\end{keywords}

\section{INTRODUCTION}
\subsection{Motivation}
\label{subsec::motivation}
\begin{figure}[t]
    \centering
    \includegraphics[scale = 0.30]{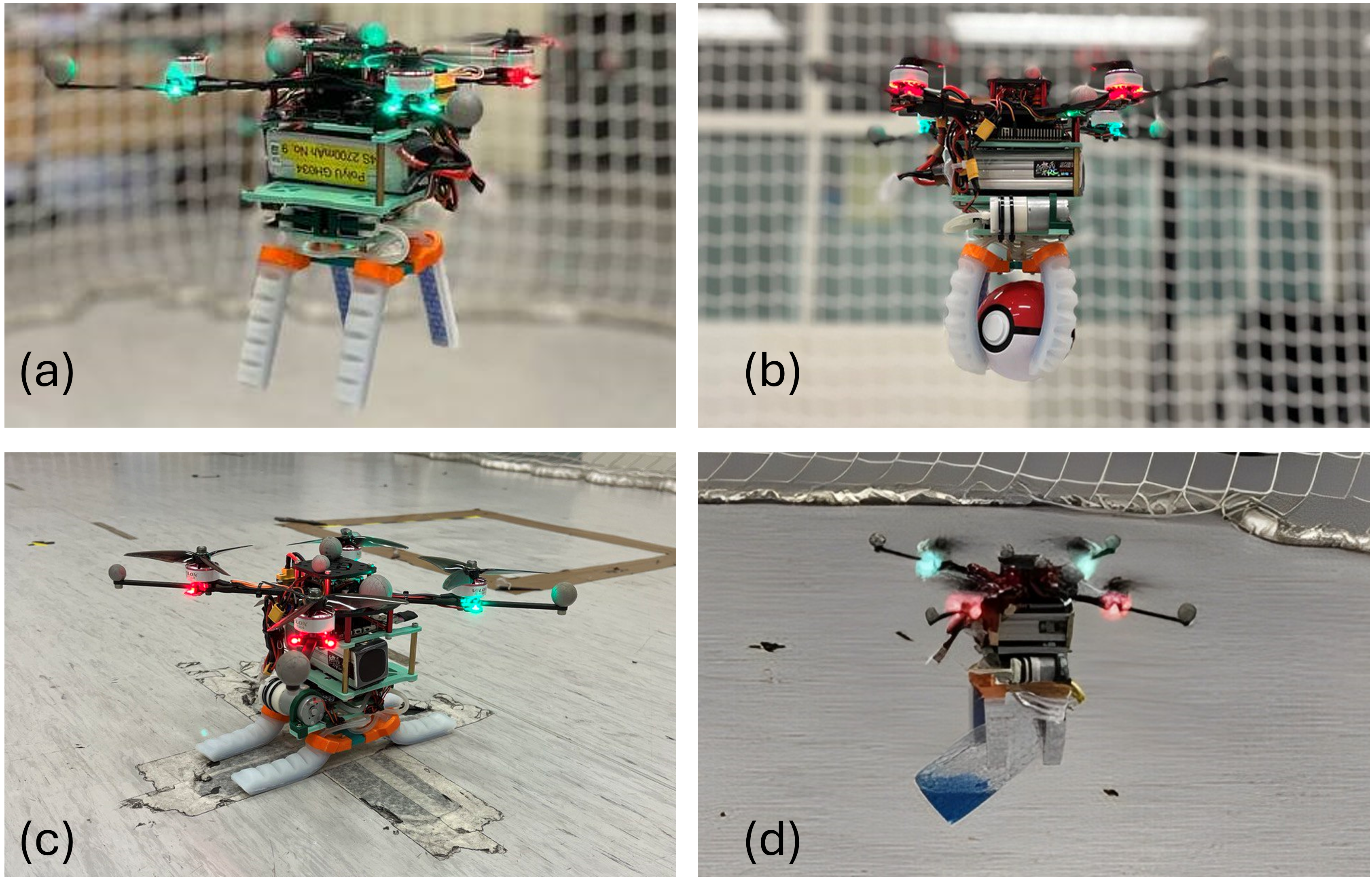}
    \caption{(a) Our SAV hovers in mid-air. (b) The SAV grasps its target object under its center of gravity by inflating its modular pneumatic soft gripper. (c) Takeoff and landing pose of the SAV with its deflated soft gripper. (d) Demonstrating the high grasping tolerance of the pneumatic soft gripper.}
    \label{fig::sav_hover}
\end{figure}

Unmanned aerial vehicles (UAVs) have gained significant attention for their diverse potential applications \cite{8299552, RADOGLOUGRAMMATIKIS2020107148, otto2018optimization}. Aerial grasping, including soft aerial grasping, is particularly promising for drone delivery \cite{muchiri2022review, yeong2015review, thiels2015use, qi2017design} and harvesting \cite{kumar2024design}. However, UAV dynamics and the control of soft grippers with infinite degrees of freedom (DoF) pose significant challenges. The increased mass associated with payload grasping affects thrust prediction, and environmental disturbances like the ground effect are often unpredictable. We addressed the infinite DoF control challenge with pressure regulation in our modular pneumatic soft gripper \cite{10521918}.

Model Predictive Control (MPC) \cite{9549626} is widely used for UAVs, providing flight stability and robustness. We aim to enhance the control of the SAV during aerial grasping by incorporating a disturbance observer into our Nonlinear Model Predictive Control (NMPC) SAV controller. This approach addresses dynamic changes, additional payload effects, and unpredictable disturbances. By integrating the proposed disturbance observer into our NMPC SAV controller \cite{jiang2022neural, li2018development}, we compensate for dynamic model idealization and handle uncertainties. Our approach combines a disturbance observer-based NMPC with the SAV controller, minimizing tracking errors along all three axes and enabling precise aerial grasping.
\subsection{Related Work}
\label{sec::relatedwork}
Some existing works focus on UAV control for soft aerial perching or grasping with static payloads. Ramon et al. \cite{ramon2019autonomous} designed a soft landing gear system that enables autonomous perching of a drone on pipes for inspection and maintenance. They use a neural network to estimate the position, which is then filtered by an Extended Kalman Filter (EKF) and used as a reference for the control loop. A PID controller generates speed commands for the flight controller to regulate the drone's location. The visual estimation is also utilized for controlling the orientation of the drone.
Fishman et al. \cite{fishman2021control, fishman2021dynamic} propose a tendon-actuated soft gripper for indoor dynamic grasping. Similar to our work, their approach eliminates the need for measuring the gripper state. However, they operate the soft gripper in an open-loop manner to grasp objects with unknown shapes on unknown surfaces, relying on the centroid of the objects. They combine an adaptive geometric controller with a minimum-snap trajectory optimizer to compensate for disturbances caused by the target object and unmodeled aerodynamic effects during dynamic grasping. They observed a phenomenon called "thrusting stealing," where the grasped item may block the airflow of the propellers. The stability of the drone differs depending on the surface area of the object, while the mass remains unchanged.
Building upon \cite{fishman2021dynamic}, Ubellacker et al. \cite{ubellacker2023aggressive} present another approach for aerial grasping with onboard perception using a soft drone. Their adaptive controller only considers external disturbances for translational dynamics, instead of rotational dynamics.
Ping et al. \cite{pingaerial} integrate a soft pneumatic gripper beneath a conventional drone, similar to our design. Their soft drone utilizes a PID controller within the flight controller unit (FCU). However, their primary focus does not revolve explicitly around autonomous aerial grasping tasks or alleviating the need for additional rigid landing gear. The lack of additional degrees of freedom (DOF) in their gripper design, which could facilitate grasping, limits the grasping altitude of the drone due to the presence of rigid landing gear. They claim that their soft gripper has minimal impact on the dynamics of the drone and tested its hovering ability with various payloads weighing up to 100 g.
In \cite{sarkar2022development}, a two DOF robot arm with a pneumatic soft gripper is installed under a traditional UAV with rigid landing gear. Hence, extra mechanisms and control algorithms are needed to ensure the manipulator can reach its home and pick positions. The control of this drone relies on the Ardupilot Firmware and QGroundControl station, while the gripper requires an additional built-in controller software (Arbotix) to control the actuators.

Recently, several research studies have focused on MPC with disturbance rejection \cite{9019022} and disturbance observer \cite{YAN202335, zhang2020tracking, hu2024disturbance} for UAVs or unmanned underwater vehicles. \cite{hanover2021performance} mentioned the importance of adaptive MPC to compensate for model mismatch and prevent degraded performance. Additionally, \cite{kumar2023thrust} proposed a quadrotor control system for aerial grasping with a rigid gripper, where they suggested fusing Inertial Measurement Unit (IMU) data using an EKF to provide acceleration feedback.
\subsection{Contribution}
\label{subsec::projectplanning}
Considering the external disturbances due to the unknown weights of payloads and other unpredictable disturbances, the SAV operated by an NMPC controller based on a disturbance observer is proposed for aerial grasping. Using an EKF, the suggested disturbance observer can adapt to dynamic model changes throughout the entire aerial grasping task. The suggested aerial grasping system provides a satisfactory payload-to-weight ratio in comparison to previous works. The proposed SAV represents a significant advancement by combining lightweight modular construction with the capability for aerial grasping. 

\begin{itemize}
    \item The proposed Disturbance Observer-based Nonlinear Model Predictive Control (DOMPC) for soft aerial grasping compensates for dynamic changes caused by payload weight and other uncertainties. By utilizing the EKF to fuse IMU data from the FCU and incorporating position and velocity information for validation, the estimated linear acceleration disturbance can be integrated into our existing NMPC for improved control of the SAV.
    \item The proposed SAV equipped with the DOMPC is capable of handling both static and non-static payloads during aerial grasping. Grasping non-static payloads enhances the applicability of our SAV, enabling it to contribute effectively to tasks such as drink delivery or environmental cleaning.
    \item The proposed lightweight soft gripper, coupled with a traditional quadrotor weighing 1002 g, successfully grasps an analytical Weight Box weighing 279 g while in mid-air. Furthermore, we demonstrate the aerial grasping of a 159 g spherical container. The payload-to-weight ratio achieved by our SAV surpasses those observed in previous investigations on soft grasping \cite{10521918, fishman2021control, pingaerial, sarkar2022development}.
\end{itemize}

\section{Soft Aerial Vehicle Design}
\label{sec::savdesign}
\begin{figure}[t]
    \centering
    \includegraphics[scale = 0.22]{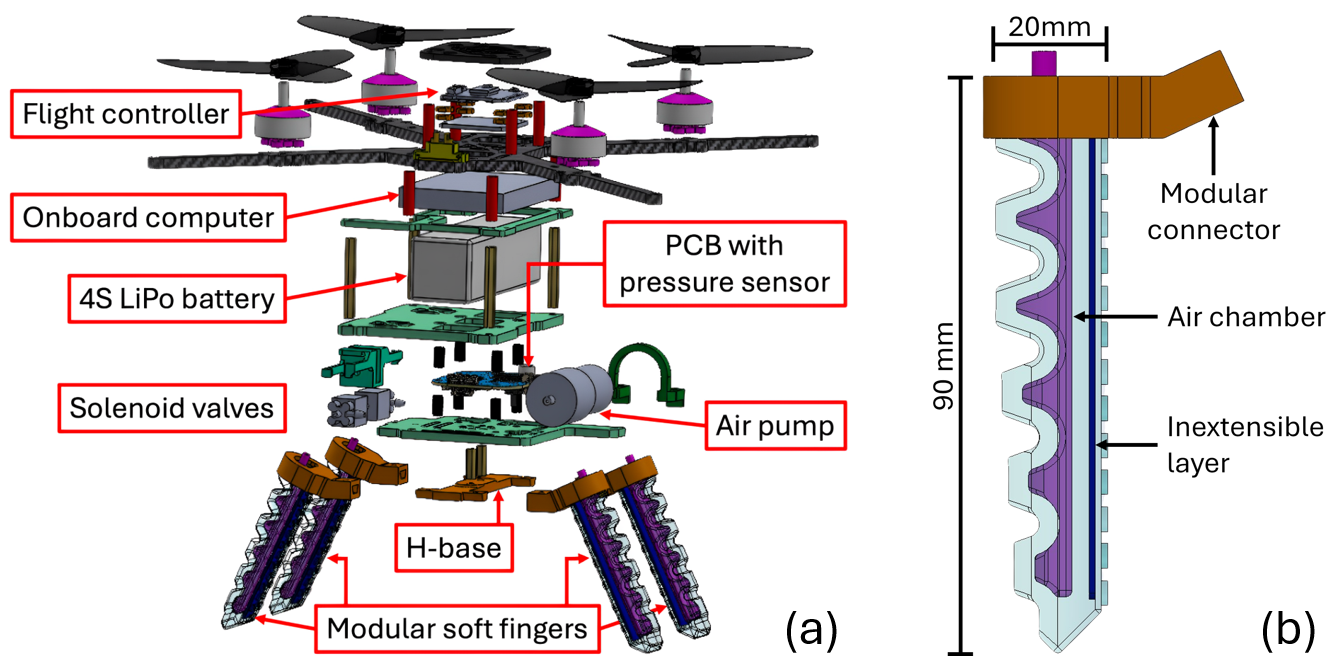}
    \caption{(a) Exploded CAD view of the SAV. (b) Side view of a modular pneumatic soft finger.}
    \label{fig::sav_explode}
\end{figure}

\subsection{Overall System Design}
\label{subsec::systemoverview}
All main components of the proposed SAV are depicted in Fig. \ref{fig::sav_explode}(a). The flight controller, Holybro Kakute H7 v1.3, is capable of interacting with the onboard computer, Khadas VIM4, via the Robot Operating System (ROS), thereby improving the real-time control efficiency of the SAV. Moreover, the quadrotor base is a customized carbon fiber board fabricated by CNC milling since the carbon fiber composites can offer a high strength-to-weight ratio to increase the durability of the SAV. The remaining customized connecting boards are manufactured through 3D printing with Polylactic Acid (PLA) due to its lightweight properties. The SAV has a diagonal length of 385 mm and a height of 245 mm. The net weight of the quadrotor base of the SAV is 732 g, and the weight of the soft gripper remains the same in both configurations.

\begin{figure}[t]
    \centering
    \includegraphics[scale = 0.30]{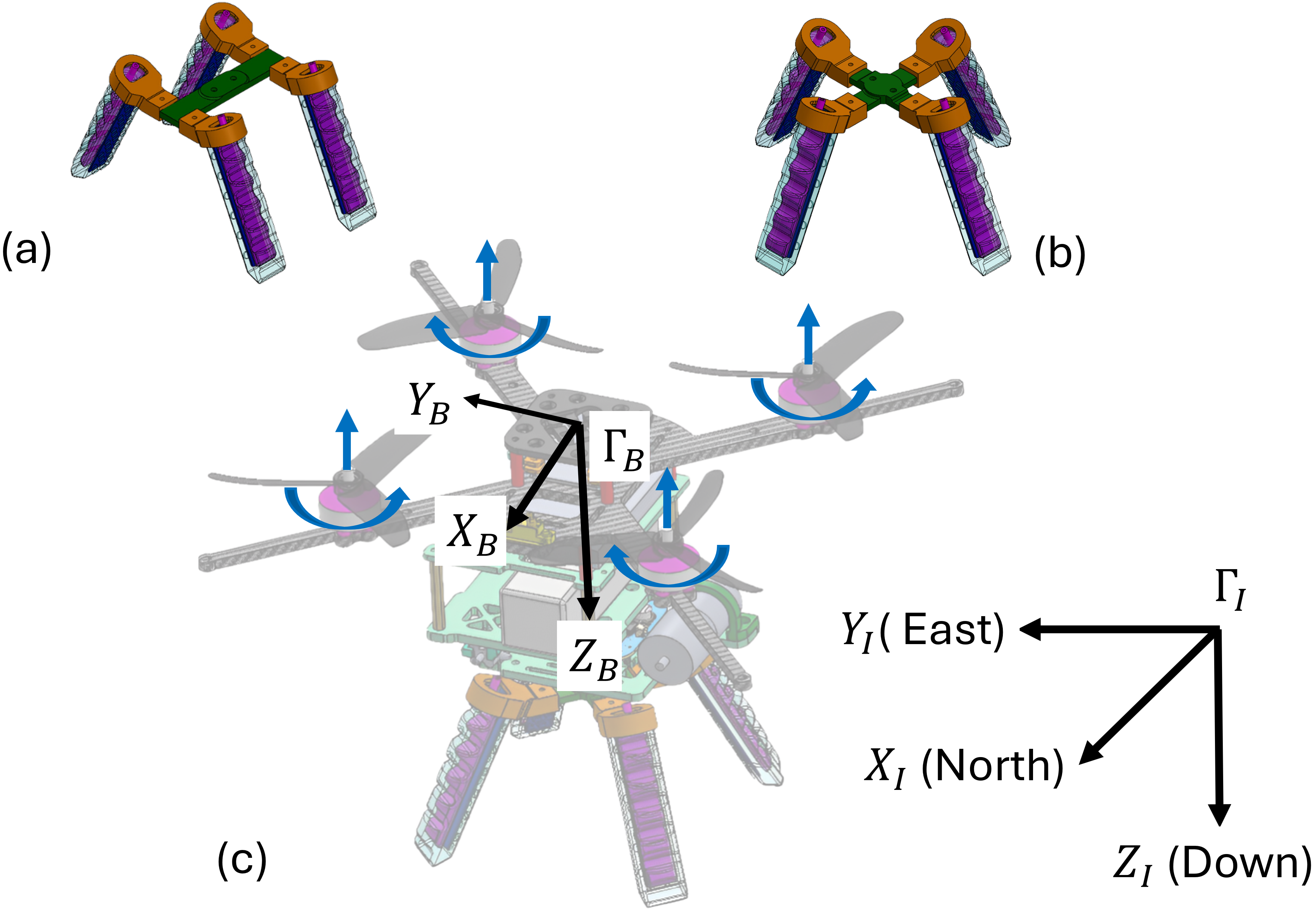}
    \caption{(a) Assembly of H-base (cylindrical) soft gripper. (b) Assembly of X-base (spherical) soft gripper. (c) SAV sketch with its inertial frame $\Gamma_I$ and body frame $\Gamma_B$.}
    \label{fig::sav_dynamics}
\end{figure}

To simplify the grasping motion and overcome the complexity of controlling the countless degrees of freedom in soft robotics, a pneumatic soft gripper is equipped with four soft fingers that can be inflated or deflated simultaneously \cite{10521918} by low-level feed-forward proportional control of pressure regulation. Hence, this proposed soft gripper is placed under the center of gravity of the quadrotor and can grasp various target objects by controlling the inflation and deflation of its fingers. Moreover, the gripper can be configured in two ways (spherical and cylindrical in Fig. \ref{fig::sav_dynamics}(a) - (b)), providing flexible options for gripping target objects based on their geometries. 

\subsection{Modular Soft Gripper Design}
\label{subsec::modularsoftfingerdesign}
The electronics design of the modular soft gripper system is the same as in our previous work \cite{10521918}. However, we have made improvements to the mechanical structure of the soft fingers to enhance grasping performance. Fig. \ref{fig::sav_explode}(b) illustrates the structure of a soft finger with dimensions. Our soft fingers have dimensions of 90 mm in length and 20 mm in width. These fingers are attached to the X-base (spherical) and the H-base (cylindrical) at an inclination angle of 25$^{\circ}$. Grasping tolerance refers to the available dimension for grasping when the gripper is deflated, specifically the distance between the fingertips when the gripper is fully open. The grasping tolerance for the H-base soft gripper is 150 x 90 mm$^{2}$, while the x-base soft gripper has a grasping tolerance of 160 mm diagonal distance. The soft fingers are molded using \textit{Smooth-On Dragon Skin 30} silicone rubber, while all the molds, soft finger connectors, and bases are 3D-printed using PLA \cite{wang2017prestressed}. Compared with our previous work \cite{10521918}, an inextensible layer made of Thermoplastic polyurethane (TPU) with a shore hardness of 80 A has been added to each soft finger to enhance the pinching force \cite{zhao2023palm,yu2022modeling}. 

\section{Aerial Grasping Control}
\label{sec::soft_gripper_design_and_control}
\subsection{System Dynamics}
\label{subsec::systemdynamics}\

To simplify the dynamics of the SAV, the inflation and deflation of the soft gripper are neglected. Considering the characteristics of a conventional drone, we assume that the SAV can be treated as a rigid body with six degrees of freedom. Let $\mathbf{X}$ represent the system state vector of the SAV, and ${\mathbf{U}}$ denote the four control inputs for the SAV. The equations governing the system state vector and the control inputs are as follows:

\begin{equation}\label{eqn::systemstateerror}
    \mathbf{X}= \left [ \mathbf{P},\ \mathbf{V},\ \Theta,\ \omega \right ]^{T}
\end{equation}

\begin{equation}\label{eqn::4controlinputs}
    \mathbf{U}=[T,\ \tau_x,\ \tau_y,\ \tau_z]
\end{equation}

\noindent where ${\mathbf{P} = [x,\ y,\ z]}, {\mathbf{V} = [u,\ v,\ w]}, {\mathbf{P}, \mathbf{V}} \in \mathbb{R}^3$ represent position and velocity in the North-East-Down inertial frame, $\Gamma_I$. And $\Theta = [\phi,\ \theta,\ \psi]$ denote Euler angles in roll, pitch, and yaw axes, while $\omega = [p,\ q,\ r]$ denotes angular rates in the body frame, $\Gamma_B$, respectively.
For the control inputs, $T$ is the thrust force along the $Z_B$ direction, and $\tau_x$, $\tau_y$, and $\tau_z$ are the rolling, pitching, and yawing moments in $\Gamma_B$, respectively. 

However, after grasping, the mass of the target objects will cause the mass changes of the SAV due to the increase in total mass. Several external disturbances that cannot be estimated precisely, such as the ground effect and the thrust dearth due to battery consumption. Therefore, we add the predicted linear acceleration disturbances in the dynamic of the SAV to overcome these external disturbances by our disturbance observer. Based on Newton-Euler formalism \cite{bresciani2008modelling}, external linear acceleration disturbances $\mathbf{d_{W}} = [\mathbf{d_{W_x}},\ \mathbf{d_{W_y}},\ \mathbf{d_{W_z}}] \in \mathbb{R}^3$ in $\Gamma_I$ are defined in the following linear acceleration equations in the SAV dynamic:

\begin{equation}
	\label{eqn::updateddynamics}
    \left\{
	\begin{aligned}
        \dot{\mathbf{P}} &= \mathbf{V} \\
        \dot{\mathbf{V}} &= R(\Theta)*(\mathbf{T}-\mathbf{g})+\mathbf{d_{W}} \\
        \dot{\Theta} &= R(\Theta)*\omega \\
        \dot{\omega} &= \frac{\tau - \omega \times (\mathbf{I} * \omega) }{\mathbf{I}}\\
	\end{aligned}
    \right.
\end{equation}
 
\noindent where $R \in SO3$ is the rotation matrix, $m$ is the mass of the SAV, $\mathbf{g}$ is the gravity force vector, $\mathbf{T} = [0, 0, T]$, $\tau = [\tau_x,\ \tau_y,\ \tau_z] \in \mathbb{R}^3$, and $I = [I_x,\ I_y,\ I_z] \in \mathbb{R}^3$ is the moment of inertia of the SAV.

\subsection{Disturbance Observer Design}
\label{subsec::disturbanceobserver}
To ensure the comprehensiveness of the disturbance observer, the observer is designed to mitigate both expected and unexpected disturbances. This includes accounting for factors such as the mass of the target object while holding it in mid-air, environmental disturbances, and measurement noises. By incorporating the EKF to construct a disturbance observer \cite{hu2024disturbance}, it becomes possible to effectively estimate disturbances on sensor measurements, including position ${\mathbf{P}}$ obtained from a motion capture system, linear velocity ${\mathbf{V}}$ from the FCU, and linear acceleration $\mathbf{\tilde{a}_{B}} = [\tilde{u_{B}},\ \tilde{v_{B}},\  \tilde{w_{B}}] \in \mathbb{R}^3$ from the inertial measurement unit (IMU) in the FCU. The system states $\chi$ and measurement states $\zeta$ of the SAV can be presented as: 

\begin{equation}
    \chi = [\mathbf{P},\ \mathbf{V},\ \mathbf{\hat{d}_{B}}]^{T}
\end{equation}

\begin{equation}
    \zeta = [\mathbf{P},\ \mathbf{V},\ \mathbf{\tilde{a}_{B}}]^{T}
\end{equation}
\noindent where $\hat{d}_{B} = [\hat{d}_{B_x},\ \hat{d}_{B_y},\ \hat{d}_{B_z}] \in \mathbb{R}^3$ denotes the predicted linear acceleration disturbance in the body frame of the SAV.

Hence, the system dynamics function is defined as follow: 
\begin{equation}
    f(\chi, u) = [\mathbf{V},\ \mathbf{\tilde{a}_{B}},\ \mathbf{\hat{d}_{B}}]^{T}
\end{equation}

Note that the estimation of the linear acceleration disturbance varies inversely with the noise present in the IMU. In other words, as the noise in the IMU decreases, the accuracy of the estimated linear acceleration disturbance improves. This highlights the importance of having a reliable and accurate IMU to enhance the disturbance estimation process and subsequently improve the overall performance of the aerial grasping system. Thus, the process noise covariance matrix $\mathbf{Q}$ and the measurement noise covariance matrix $\mathbf{R}$ are tuned to get a satisfied disturbance estimation. The two covariances are computed as follows: 

\begin{equation}
    Q = \text{diag} [Q_P\ Q_V\ Q_{\hat{d}}] = \text{diag} \left[\frac{({\Delta t})^{q_p}}{q_p}\ \frac{({\Delta t})^{q_v}}{q_v}\ ({\Delta t})^{q_d}\right]
\end{equation}

\begin{equation}
    R = \text{diag} [R_P\ R_V\ R_{\tilde{a}}] = \text{diag} \left[({\Delta t})^{r_p}\ ({\Delta t})^{r_v}\ \frac{({\Delta t})^{r_{\tilde{a}}}}{r_{\tilde{a}}}\right]
\end{equation}
\noindent where $\Delta t$ is the sample time step, which is 0.01 second, $q_p = q_v =r_{\tilde{a}} = 2$, $r_p = r_v = 4$, $q_d = [q_{d_x},\ q_{d_y},\ q_{d_z}] = [4.2,\ 4.2,\ 3.5]$. By our observation, the position and velocity measurements of the SAV from the indoor motion capture system and the FCU are reliable and not noisy. However, the linear acceleration data from IMU is quite noisy. As a result, we trust the predicted linear acceleration results more than those from the measurement model, while we trust the measured position and velocity results more than those from the prediction model.

In the prediction step, we compute the Jacobian of the system dynamics and predict the next state using a 4th-order Runge-Kutta (RK4) method as: 

\begin{equation}
    \begin{aligned}
        \label{eqn::RK4}
        \chi_{k+1} &= f(\chi, \mathbf{\hat{d}_{B}}, k) = \chi_{k} + \frac{1}{6}(K_1 + 2K_2 + 2K_3 + K_4) \\
        K_1 &= \Delta k \cdot {f}(\chi_{k}, \mathbf{\hat{d}_{B}}) \\
        K_2 &= \Delta k \cdot {f}\left(\chi_{k} + \frac{K_1}{2}, \mathbf{\hat{d}_{B}}\right) \\
        K_3 &= \Delta k \cdot {f}\left(\chi_{k} + \frac{K_2}{2}, \mathbf{\hat{d}_{B}}\right) \\
        K_4 &= \Delta k \cdot {f}\left(\chi_{k} + K_3, \mathbf{\hat{d}_{B}}\right)
    \end{aligned}
\end{equation}

The Jacobian of the system dynamics $\mathbb{F}$ and covariance prediction $\mathbb{P}_{k-1|k}$ are computed in Equations \ref{eqn::jacobiansystemdynmaics} and \ref{eqn::covarianceprediction}.
\begin{equation}\label{eqn::jacobiansystemdynmaics}
    \mathbb{F}_k = \frac{\partial f}{\partial \chi}\bigg|_{\chi_k, u_k}
\end{equation}

\begin{equation}\label{eqn::covarianceprediction}
    \mathbb{P}_{k|k-1} = \mathbb{F}\mathbb{P}_{k}\mathbb{F}^{T}+Q
\end{equation}

For the update step, we define the measurement model function $\Upsilon$ as follows: 
\begin{equation}
    \Upsilon = h(\chi_k) = [\mathbf{P},\ \mathbf{V},\ \mathbf{\tilde{a}_{B_{t_{m}}}}]^{T} 
\end{equation}
where $\tilde{a}_{B_{t_{m}}} = [\tilde{u_{B}},\ \tilde{v_{B}},\  \tilde{w_{B}}+ t_{m}]$ and $t_{m}$ is the thrust-to-hover thrust ratio multiplied by gravity. Since we are only focusing on the effects of uncertainties, we add $t_m$ into the measurement model function to exclude the linear acceleration caused by gravity.

The Jacobian of measurement model $\mathbb{H}$, and measurement residual $\hat{y}_{k|k}$ are formulated as the Equations \ref{eqn::jacobianofmeasurementmodel} and \ref{eqn::measurementresidual}, where $z_k$ is the actual measurement.

\begin{equation}\label{eqn::jacobianofmeasurementmodel}
    \mathbb{H}_k = \frac{\partial h}{\partial \chi}\bigg|_{\chi_{k|k-1}}
\end{equation}

\begin{equation}\label{eqn::measurementresidual}
    \hat{y}_{k|k} = z_k - h(\hat{\chi}_{k|k-1})
\end{equation}

Then the Kalman Gain $\mathbb{K}$ at time k can be computed according to $\mathbb{P}_{k|k-1}$ and $\mathbb{H}_k$. Eventually, the state estimate $\hat{\chi}_{k|k}$ can be updated by the predicted state estimate $\hat{\chi}_{k|k-1}$ and $\mathbb{K}$, and then recalculate $\mathbb{P}_{k|k}$ by ${K}_k$ and $\mathbb{H}_k$. 

\begin{equation}
    \begin{aligned}
        \label{eqn::EKF}
        K_k &= \mathbb{P}_{k|k-1}\mathbb{H}_k^{T}(\mathbb{H}_k \mathbb{P}_{k|k-1}\mathbb{H}_k^{T} + R_k)^{-1}\\
        \hat{\chi}_{k|k} &= \hat{x}_{k|k-1} + {K}_k \hat{y}_{k|k} \\
        \mathbb{P}_{k|k} &= (I - {K}_k \mathbb{H}_k)\mathbb{P}_{k|k-1} \\
    \end{aligned}
\end{equation}

\subsection{Disturbance Observer-Based Nonlinear Model Predictive Control}
\label{subsec::savcontrol}

\begin{figure}[t]
    \centering
    \includegraphics[scale = 0.18]{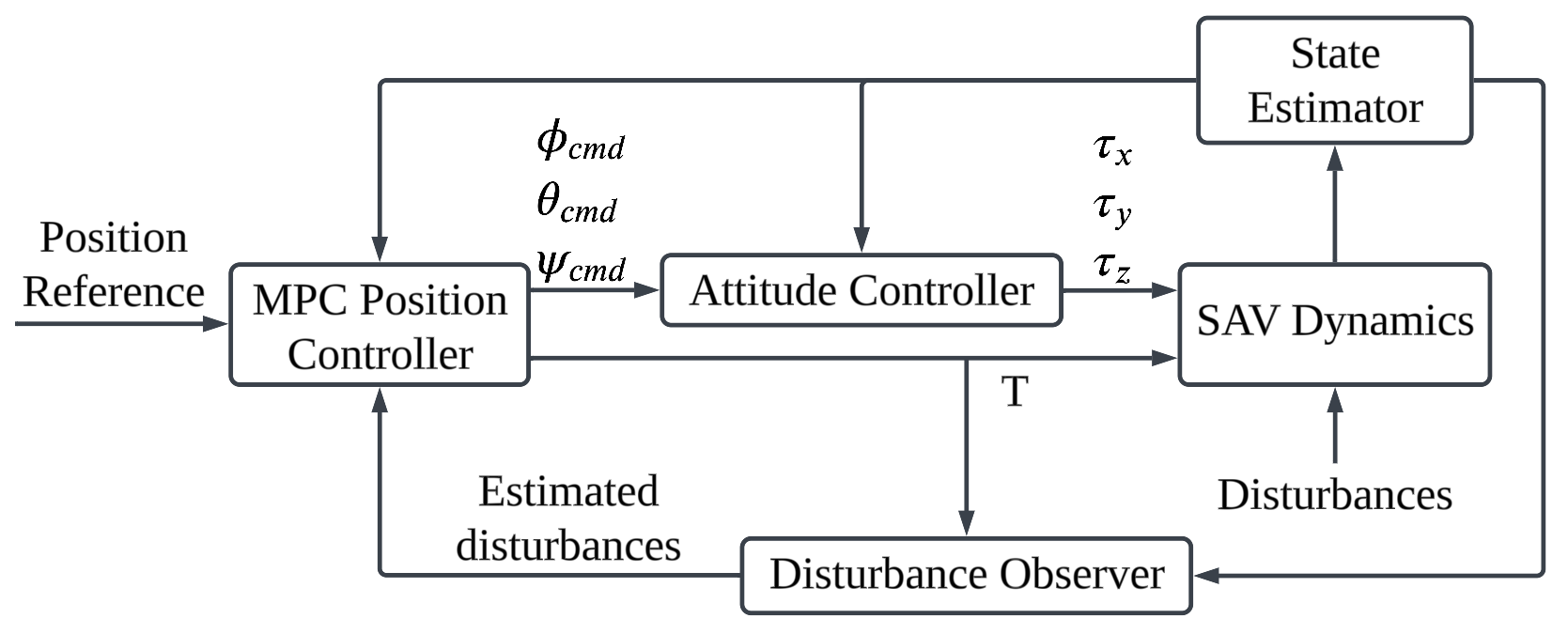}
    \caption{Cascaded loop control structure of disturbance observer-based NMPC (DOMPC).}
    \label{fig::sav_systemcontrol}
\end{figure}

\begin{figure}[t]
    \centering
    \includegraphics[scale = 0.20]{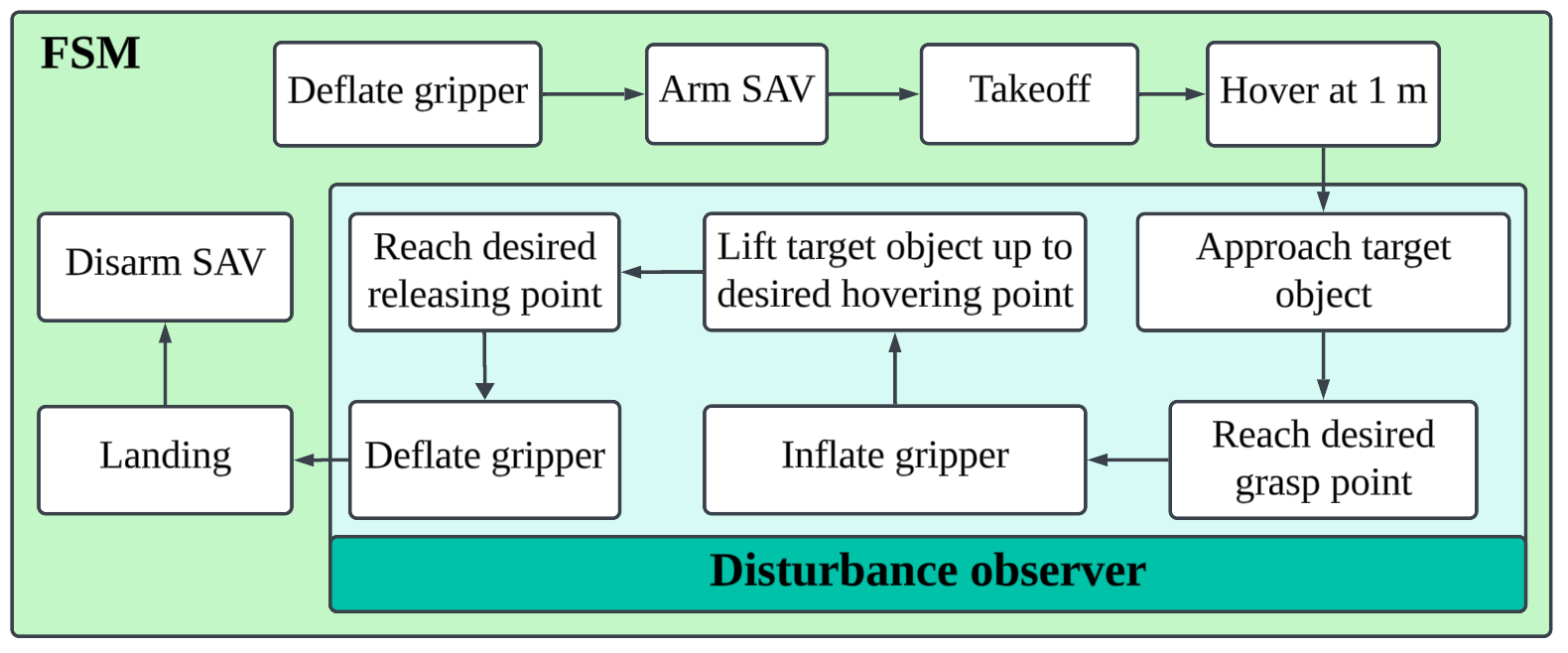}
    \caption{Finite state machine diagram for the SAV aerial grasping task.}
    \label{fig::sav_fsm}
\end{figure}

After the SAV grasps its target object, the total mass of the SAV increases due to the addition of the mass of the target object. 
Consequently, the changes in the dynamic model of the SAV potentially result in degraded performance of the NMPC. However, this degradation can be mitigated by incorporating a disturbance observer in the NMPC. The proposed disturbance observer, described in section \ref{subsec::disturbanceobserver}, effectively handles changes in dynamics that arise from varying operational circumstances. To account for these changes, linear acceleration disturbances are added to the linear acceleration equations of the SAV's dynamic model in Equation (\ref{eqn::updateddynamics}) and the nonlinear prediction model in its NMPC controller in Equation (\ref{eqn::quadrotor_mpc_model}). These disturbances are transformed from $\Gamma_B$ to $\Gamma_I$ by multiplying the rotation matrix. By incorporating the disturbance observer, the NMPC system can accurately estimate and compensate for these disturbances, thereby enhancing the overall control performance of the SAV. The cascaded control structure of the DOMPC is described in Fig. \ref{fig::sav_systemcontrol}, while the finite state machine diagram is illustrated in Fig. \ref{fig::sav_fsm}.

The nonlinear prediction model in the NMPC controller is formulated as \cite{jiang2022neural}: 

\begin{equation}\label{eqn::quadrotor_mpc_model}
	\left\{
	\begin{aligned}
        \dot{\mathbf{P}} &= \mathbf{V} \\
        \dot{\mathbf{V}} &= R(\Theta)*(\mathbf{T}-g)+\mathbf{d_{W}} \\
		\dot{\phi}&=\frac{\phi_{cmd}-\phi}{\tau_\phi}\\
		\dot{\theta}&=\frac{\theta_{cmd}-\theta}{\tau_\theta}
	\end{aligned}
	\right.
\end{equation}
            
\noindent where $\tau_\phi$ and $\tau_\theta$ are time constants of roll and pitch control, $\phi_{cmd}$ and $\theta_{cmd}$ are roll and pitch commands sent to inner loop attitude control. The yaw angle is not included in the above NMPC states as the inner loop controller does the yaw angle control. The system identification technique can be used with flight data to derive the values of $\tau_\phi$ and $\tau_\theta$. 


The estimated disturbances obtained from the disturbance observer can be effectively incorporated into the prediction horizon $N$ of the NMPC algorithm and updated at each time step. By considering the estimated disturbances during the control optimization process, the NMPC can generate optimal control inputs that effectively reject disturbances and enhance the overall performance of the aerial grasping system. 

The optimizer solves the Quadratic Programming (QP) problem formulated as \cite{jiang2022neural}:

\begin{equation}\label{eqn::quadrotor_cost_funciton}
	\begin{aligned}
		\min \quad &\int_{t=0}^{N} ||h(x(t),u(t))-y_{ref} ||_Q^2 dt\\
		&\quad \quad + || h(x(T))-y_{N,ref} ||_{Q_N}^2 dt
		\\
		s.t.\quad &\dot{x}=f(x(t),u(t))\\
		&u(t)\in \mathbb{U}\\
		&x(t)\in \mathbb{X}\\
		&x(0)=x(t_0),\\
	\end{aligned}
\end{equation}

\noindent where $u(t)$ and $x(t)$ represent the control input and state at timestep $t$, respectively. $y_{ref}$ and $y_{N,ref}$ denote the reference state for the prediction horizon and terminal timestep, respectively. The weighting matrices for states and terminal states are represented by $Q$ and $Q_N$, while $f(\cdot)$ and $h(\cdot)$ indicate the prediction function and system output function, respectively. $\mathbb{U}$ and $\mathbb{X}$ are the input constraint and state constraint.

The Optimal Control Problem (OCP) in Eqt. (\ref{eqn::quadrotor_cost_funciton}) is solved using the Multiple Shooting Method with the Active Set Method and qpOASES solver, employing the Sequential Quadratic Programming (SQP) technique \cite{Kamel2017}. Real-time computation is achieved through the implementation of NMPC with the "acados" solver, which offers efficient solutions for estimation and optimal control problems \cite{verschueren2022acados}.
\section{EXPERIMENTAL RESULTS AND DISCUSSION}
\label{sec::resultsanddiscussion}
\subsection{Experiment Implementation}
\subsubsection{Observer Performance Test}
\label{subsubsec::observertest}
To evaluate the performance of the DOMPC, the SAV carried an additional 257 g payload (a 4S LiPo battery) and followed a circular trajectory with a radius of 1.8 m at a speed of 2 m/s and an altitude of 1.0 m by DOMPC, NMPC, and PID controllers respectively. The flight tests were conducted utilizing a VICON motion capture system, which provided real-time ground truth data to assist the flight controller in tracking the designated setpoints. Not to mention, the load was positioned off-center from the SAV's center of gravity, next to the original SAV's battery, to add additional unpredictable variables to the system.

\subsubsection{SAV Payload Test}
\label{subsubsec::payloadtest}
\begin{figure}[t]
    \centering
    \includegraphics[scale = 0.24]{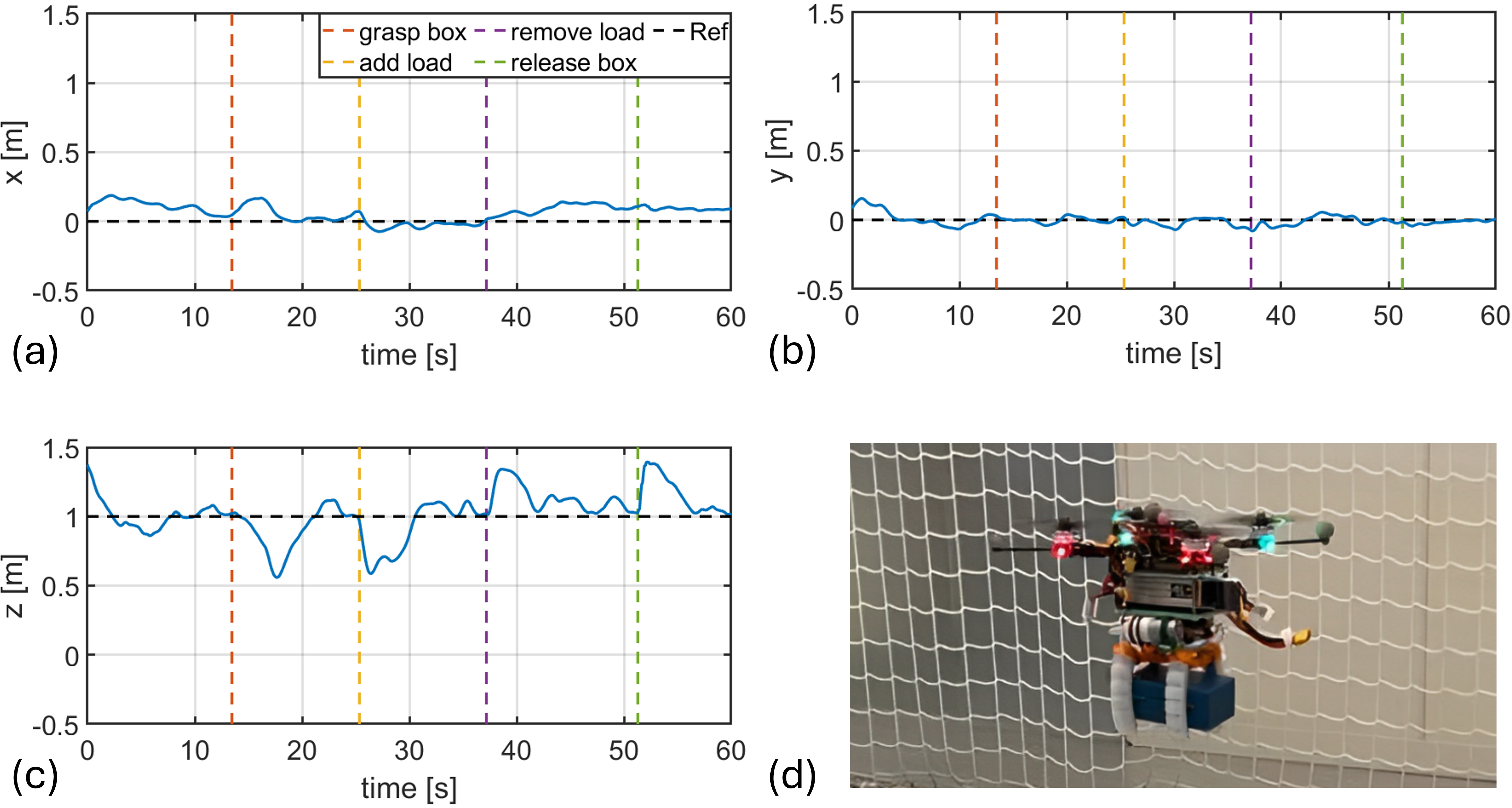}
    \caption{Payload test of the SAV: (a) - (c) position tracking performance along the x, y, and z axes, (d) SAV hovering after carrying 634 g payload in total.}
    \label{fig::payload}
\end{figure}

To ascertain the maximum payload capacity of the SAV, it not only carried the same load utilized in the observer performance test during mid-air operation, but also grasped an analytical weight box with varying weights while hovering. Following the SAV's hover (equipped with an H-base soft gripper) the analytical weight box was initially positioned beneath it for grasping. The maximum weight of the analytical weight box was 337 g. Subsequently, after the SAV grabbed the box and returned to the desired hovering point, an additional 257 g load was placed next to its battery. The box was released after the SAV reached the target hovering point with both the box and the load. Similarly, the load was removed after the SAV resumed hovering at the desired point. Figure \ref{fig::payload} illustrates the position tracking of the SAV. During this payload test, although the altitude of the SAV decreased when it began grasping the box and carrying the load, it ultimately maintained its hovering position with the proposed disturbance observer. Its hovering position was also sustained when its payload was released or removed. The resultant payload capacity (634 g) surpasses that of several existing research projects, such as a 100 g payload in \cite{pingaerial}, a 106 g foam target in \cite{fishman2021dynamic}, a 148 g med-kit in \cite{ubellacker2023aggressive}, and a 150 g packet in \cite{sarkar2022development}. The primary objective of this test was to evaluate the payload capacity of the SAV, intentionally excluding the assessment of its aerial grasping capability. A similar approach for the payload test of the SAV utilizing its X-base soft gripper is demonstrated in the accompanying video.

\subsubsection{Soft Aerial grasping Task}
\label{subsubsec::softaerialgraspingtask}
\begin{table}[t]
 \begin{center}
 \renewcommand{\arraystretch}{1.3}
 \caption{Dimensions and weights of target objects.}
        \begin{tabular}{@{}>{\centering\arraybackslash}m{3.2cm}|>{\centering\arraybackslash}m{2.9cm}|>{\centering\arraybackslash}m{1.4cm}@{}}
        \hline 
            Target objects & Dimensions/Volumes & Weights (g) \\
		\hline 
		  Shuttlecock Tube with 46 g load & 238 mm*(33 mm)$^2$$\pi$ & 113 \\
          Spherical container with 118 g load & $\dfrac{4}{3}$$\pi$*(35 mm)$^3$ & 161\\ 
          Plastic bottle with 80 g dyed water & 210 mm*60 mm*60 mm & 110\\
        \hline
	\end{tabular}
    \renewcommand{\arraystretch}{1}
	\label{tab::dimensionsweight}
  \end{center}
\end{table}

 In the aerial grasping task, we tested the aerial grasping capabilities  of the SAV with both static and non-static payloads. For the preparation of target objects, additional loads were placed into a spherical container, affixed to the side of the cylindrical shuttlecock tube, and dyed water was introduced into a rectangular plastic water bottle. The dimensions and weights of the three target objects can be found in Table \ref{tab::dimensionsweight}. 


The estimated disturbance would not be utilized during takeoff and landing to mitigate the ground effect, friction, and contact force experienced by the SAV. Since this work focuses on the disturbance observer's ability to estimate unknown payload and other uncertainties that can affect the dynamics of the SAV, we refrained from providing a predefined trajectory for the SAV during the aerial grasping task. Instead, we instructed the SAV to fly point-to-point manner, evaluating its ability to reach the desired points with minimal tracking errors. The soft gripper received a deflation command from the flight controller via ROS before take-off. After successfully reaching the desired grasping point, the flight controller sent a PWM command for inflation to the soft gripper's controller. The SAV remained stationary for at least 5 seconds to ensure the soft gripper was fully inflated, thereby testing its ability to overcome the ground effect. 
\subsection{Results and Discussion}
\subsubsection{Disturbance Observer Performance}
\label{subsubsec::ekf_observer_performance}

\begin{figure}[t]
    \centering
    \includegraphics[scale = 0.25]{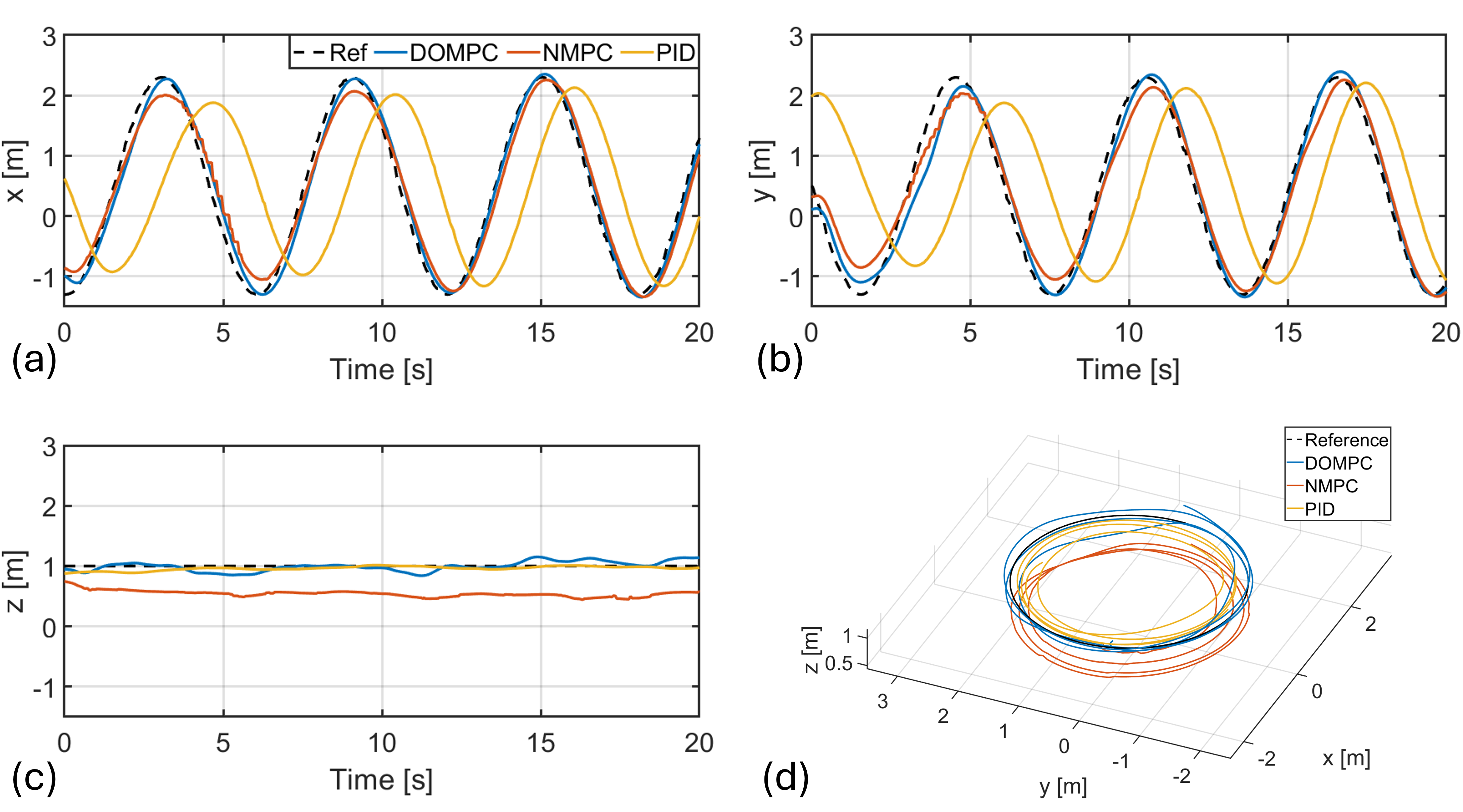}
    \caption{Results of the SAV circular flights with an additional 257g load using DOMPC, NMPC, and PID: (a) - (c): position tracking performance along the x, y, and z axes; (d): trajectories results of the SAV circular flights.}
    \label{fig::battcirtrackingerror}
\end{figure}


The resulting statistics for the position tracking performance are presented in Fig. \ref{fig::battcirtrackingerror}. The DOMPC exhibited superior tracking performance in both the x and y directions and demonstrated comparable accuracy to the PID controller in the z direction. In contrast, the NMPC controller showed significant tracking errors, particularly in the z direction as it could only maintain its altitude at around 0.5 meters. As discussed in section \ref{subsec::savcontrol}, the degraded performance of the NMPC is caused by significant changes in the dynamic model of the SAV, such as unmodelled mass addition due to its payload. Hence, the SAV's altitude tracking error with the NMPC was around 0.5 m, while the SAV with the DOMPC could track the target altitude successfully. Integrating the proposed disturbance observer into the NMPC presents an effective method for compensating the SAV's dynamic changes and ensuring accurate path tracking. However, only using the NMPC and PID controllers is less robust for handling dynamic changes due to unknown payloads.

\subsubsection{Soft Aerial Grasping Performance}
\label{subsubsec::soft_grasp_performance}

\begin{figure}[t]
    \centering
    \includegraphics[scale = 0.50]{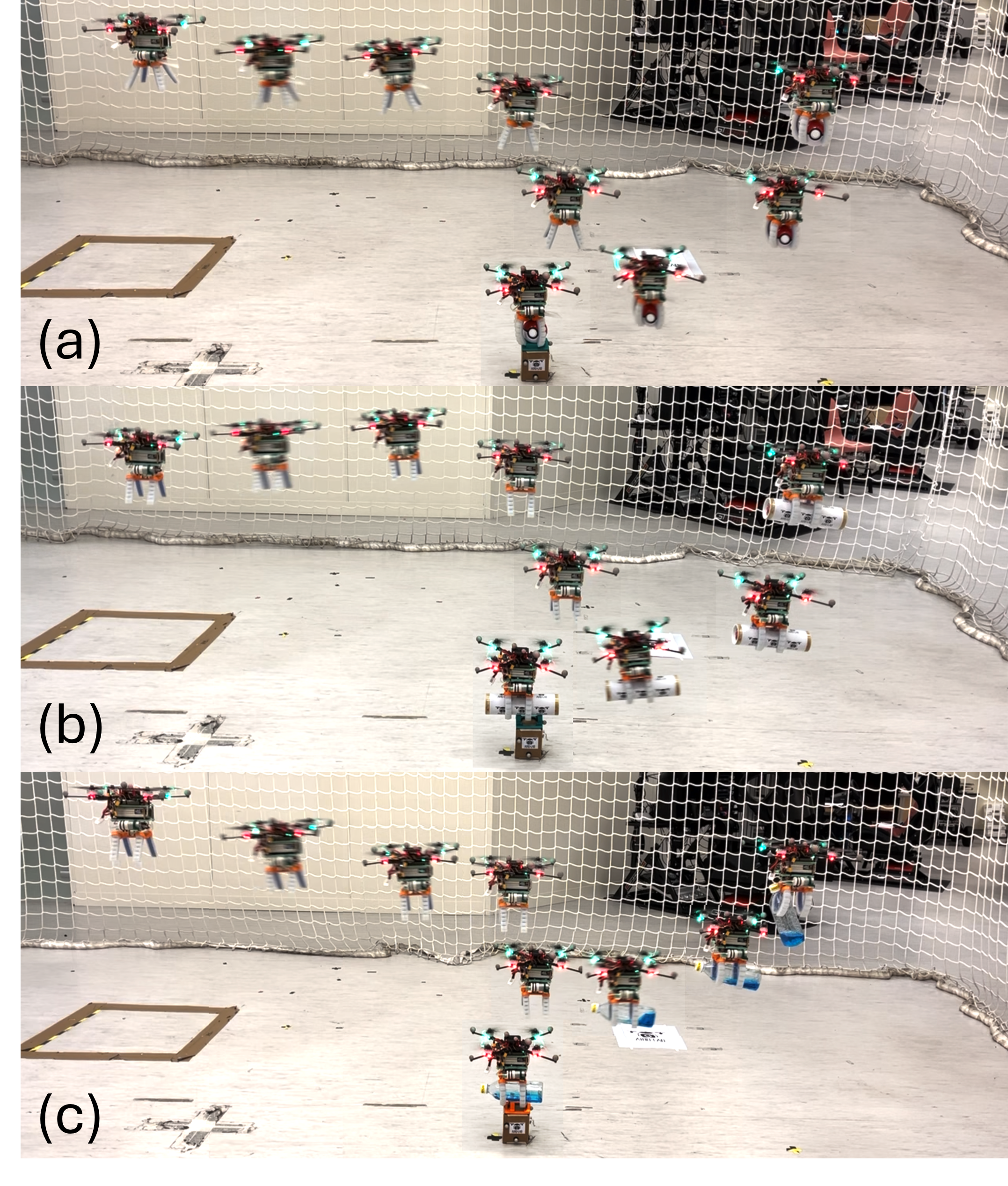}
    \caption{Trajectories of grasping the three targets: (a) grasping a shuttlecock tube with 46 g load; (b) grasping a spherical container with 118 g load; (c) grasping a plastic bottle with 80 g .}
    \label{fig::grasp_trajectory_path}
\end{figure}

\begin{figure}[t]
    \centering
    \includegraphics[scale = 0.33]{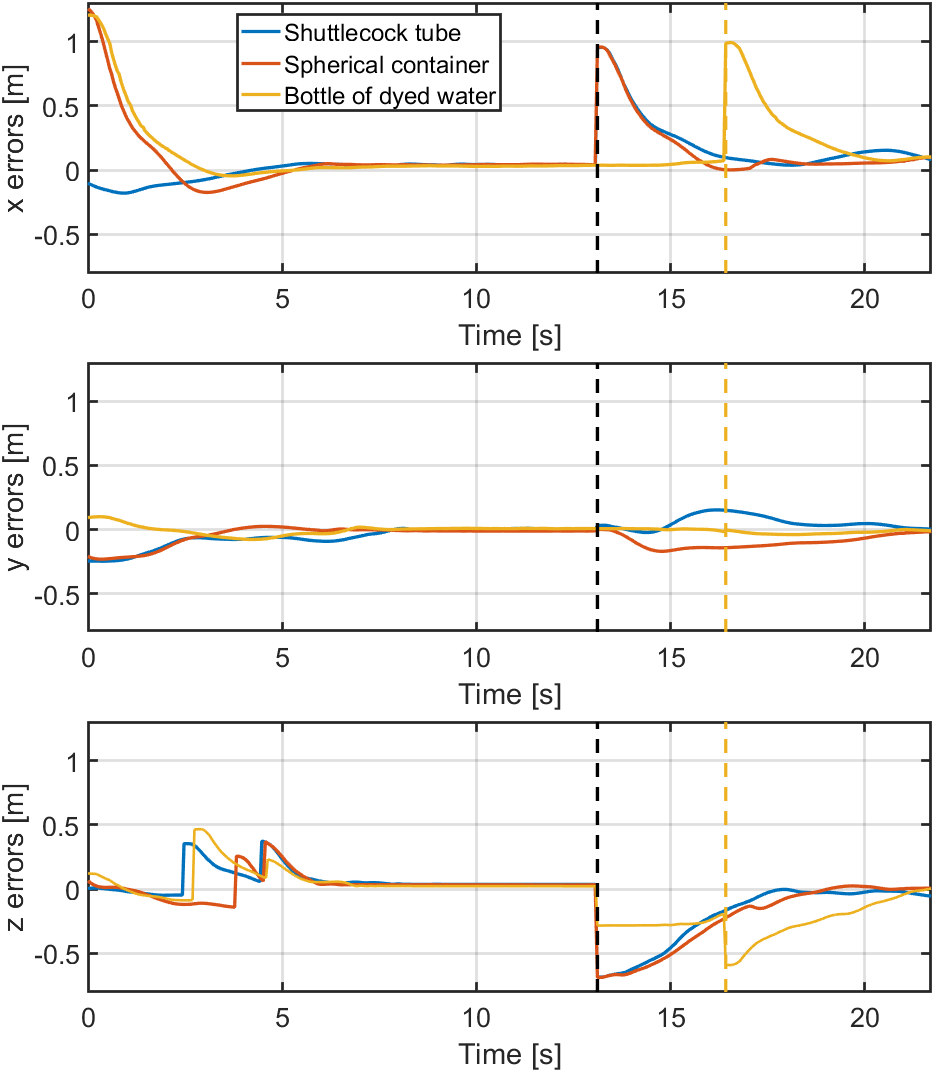}
    \caption{Position tracking errors during the process of grasping the three different target objects (the dotted lines depict the time of holding and moving the objects after grasping).}
    \label{fig::grasp_tracking_error}
\end{figure}

With the proposed disturbance observer, Fig. \ref{fig::grasp_trajectory_path} illustrates the successful aerial grasping demonstrations performed by the SAV. Equipped with DOMPC, the SAV demonstrates precise hovering capabilities over the target objects and effectively grasps them at the desired location. Fig. \ref{fig::grasp_tracking_error} shows that as the SAV was instructed to transport the target object along the x-axis at the height of 1 m following the grasping, the magnitude of tracking errors along the z-axis and x-axis rapidly increased after grasping the three target objects respectively. However, due to the disturbance detection capability of DOMPC, these errors quickly converged towards zero. The soft gripper finished inflation for grasping at 13.1 s. In contrast to the grasping of cylindrical or spherical objects, an additional setpoint was added for grasping the rectangular bottle containing non-static dyed water to ensure a secure grip. Therefore, the SAV first grasped the bottle with dyed water and hovered at a height of 0.6 m, then started flying to the desired holding location at 16.4 s. Since the SAV was instructed to fly point-to-point, the grasp time of each trial differs. The SAV took less than 22 seconds to bring the objects to the desired location due to the accuracy of the disturbance observer.

\begin{figure}[t]
    \centering
    \includegraphics[scale = 0.33]{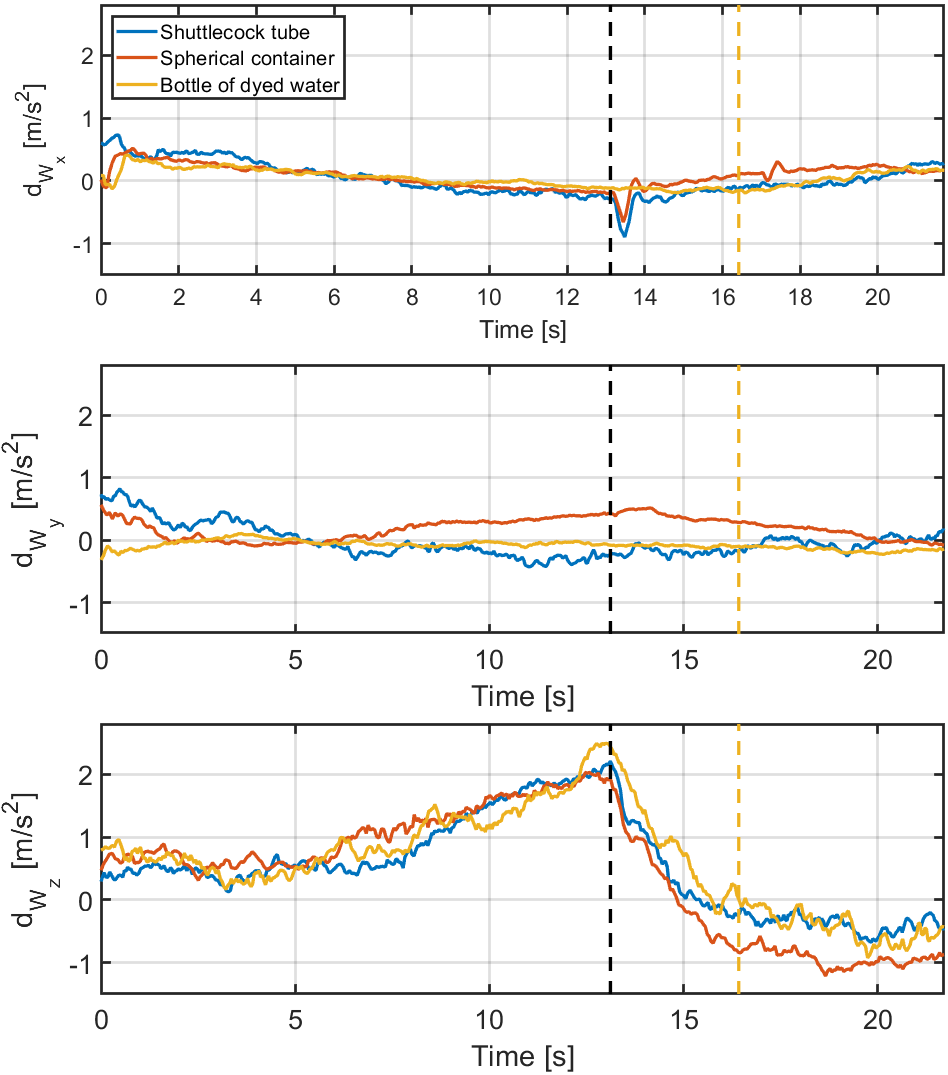}
    \caption{Disturbance results from grasping the three different target objects (the dotted lines indicate the time of holding and moving the objects after grasping).}
    \label{fig::sav_grasp_disturbances}
\end{figure}

Fig. \ref{fig::sav_grasp_disturbances} displays the estimated disturbances during the grasping process for three different objects. Notably, the spherical container (weighing 161 g) was the heaviest among the three objects, resulting in the largest estimated disturbance magnitude along the z-axis ($d_{W_z}$), approximately -1, when grasped by the SAV. For the other two objects, their magnitudes of $d_{W_z}$ are similar and slightly less than that of the spherical container due to their comparable weights (110 g and 113 g). These results indicate that the disturbance induced by the non-static mass can be estimated by the proposed observer. Besides, the dynamic changes during the SAV aerial grasping are crucially influenced by the increased mass created by the target objects, which is under the SAV. Hence, the changes in $d_{W_x}$ and $d_{W_y}$ are not as strong as those in $d_{W_z}$. These findings indicate that the proposed observer is capable of evaluating the disturbances by both static and non-static masses.

\begin{table}[t]
 \begin{center}
 \renewcommand{\arraystretch}{1.3}
 \caption{Comparison of SAV aerial grasping performance with the three target objects.}
        \begin{tabular}{@{}>{\centering\arraybackslash}m{2.0cm}|>{\centering\arraybackslash}m{3.2cm}|>{\centering\arraybackslash}m{1.7cm}@{}}
        \hline 
            Configurations & Target objects & Successes \\
		\hline 
		  H-base & Shuttlecock Tube with 46 g load & $27/30 (90\%)$ \\
          X-base & Spherical container with 118 g load & $30/30 (100\%)$ \\ 
          H-base & Plastic bottle with 80 g dyed water & $20/30 (66.7\%)$\\
        \hline
	\end{tabular}
    \renewcommand{\arraystretch}{1}
	\label{tab::successgrasp}
  \end{center}
\end{table}

The success rate of the three aerial grasping tasks is presented in Table \ref{tab::successgrasp}. Despite its relatively high weight, the spherical container with a load was successfully grasped in all trials, attributable to the soft gripper's tolerance. The grasping tolerance-to-objects' surface ratio in the spherical container grasping task was the highest. However, although the weights of the bottle containing dyed water and the shuttlecock tube containing the load were similar, the success rate for gripping the bottle was significantly lower compared to that of the shuttlecock tube. This discrepancy can be attributed to the rectangular shape of the plastic bottle, which requires higher pinching forces from the soft finger. As a result, although the SAV consistently achieved successful reachability to the bottle, it may struggle to lift it due to the limited contact area between the soft fingertips and the lateral surface of the bottle.
\section{CONCLUSION}
\label{sec::conclusion}
In conclusion, we have proposed a DOMPC system for soft aerial grasping with the SAV. By incorporating a disturbance observer and utilizing EKF, our approach effectively adapts to dynamic model changes and manages unpredictable disturbances throughout the aerial grasping task. The DOMPC framework compensates for uncertainties arising from payload weight variations and other external disturbances, such as battery discharging. 
The SAV equipped with DOMPC demonstrates the capability to handle both static and non-static payloads. It is capable of autonomously grasping an 80 g plastic bottle filled with dyed water, even when the liquid is agitated, inducing unanticipated disturbances. Furthermore, our 270 g lightweight soft gripper, combined with a 732 g customized traditional quadrotor, achieves successful mid-air grasping of various objects. This includes the analytical Weight Box (337 g) in the payload test and a spherical container (160 g) in the aerial grasping test. The payload-to-weight ratio achieved by our SAV surpasses previous investigations in the domain of soft grasping, highlighting the effectiveness of our approach. 




\bibliographystyle{IEEEtran}
\bibliography{references}

\end{document}